\pdfoutput=1
\documentclass[sigconf,nonacm]{acmart}

\usepackage{pgfplots}
\pgfplotsset{compat=newest}
\usepgfplotslibrary{groupplots}
\usepackage{tikz}
\usetikzlibrary{patterns}
\usepackage{forest}
\usepackage{tabularx}
\usepackage{booktabs}
\usepackage{makecell}
\usepackage{pgfplots}
\usepackage{hyperref}

\AtBeginDocument{%
  }

\setcopyright{acmlicensed}
\copyrightyear{2025}
\acmYear{2025}
\acmDOI{XXXXXXX.XXXXXXX}
\acmConference[ACMMM]{ACM International Conference on Multimedia}{October 27--31,
  2018}{Dublin, Irelan}
\acmISBN{978-1-4503-XXXX-X/2018/06}

\begin{document}

\title{Gaze into the Heart: A Multi-View Video Dataset for rPPG and Health Biomarkers Estimation}

\author{Konstantin Egorov}
\affiliation{%
  \institution{Sber AI Lab}
  \city{Moscow}
  \country{Russia}
}
\email{egorov.k.ser@sber.ru}

\author{Stepan Botman}
\affiliation{%
  \institution{Sber AI Lab}
  \city{Moscow}
  \country{Russia}
}
\email{SABotman@sber.ru}

\author{Pavel Blinov}
\affiliation{%
  \institution{Sber AI Lab}
  \city{Moscow}
  \country{Russia}
}
\email{Blinov.P.D@sber.ru}

\author{Galina Zubkova}
\affiliation{%
  \institution{Sber AI Lab}
  \city{Moscow}
  \country{Russia}
}
\email{GVZubkova@sber.ru}

\author{Anton Ivaschenko}
\affiliation{%
  \institution{Samara State Medical University}
  \city{Samara}
  \country{Russia}
}
\email{an.v.ivaschenko@samsmu.ru}

\author{Alexander Kolsanov}
\affiliation{%
  \institution{Samara State Medical University}
  \city{Samara}
  \country{Russia}
}
\email{a.v.kolsanov@samsmu.ru}

\author{Andrey Savchenko}
\orcid{0000-0001-6196-0564}
\affiliation{%
  \institution{Sber AI Lab}
  \institution{ISP RAS Research Center for Trusted Artificial Intelligence}
  \city{Moscow}
  \country{Russia}}
\email{avsavchenko@hse.ru}

\renewcommand{\shortauthors}{Egorov et al.}

\begin{abstract}
Progress in remote PhotoPlethysmoGraphy (rPPG) is limited by the critical issues of existing publicly available datasets: small size, privacy concerns with facial videos, and lack of diversity in conditions. The paper introduces a novel comprehensive large-scale multi-view video dataset for rPPG and health biomarkers estimation. Our dataset comprises 3600 synchronized video recordings from 600 subjects, captured under varied conditions (resting and post-exercise) using multiple consumer-grade cameras at different angles. To enable multimodal analysis of physiological states, each recording is paired with a 100 Hz PPG signal and extended health metrics, such as electrocardiogram, arterial blood pressure, biomarkers, temperature, oxygen saturation, respiratory rate, and stress level. Using this data, we train an efficient rPPG model and compare its quality with existing approaches in cross-dataset scenarios. The public release of our dataset and model should significantly speed up the progress in the development of AI medical assistants.
\end{abstract}


\begin{teaserfigure}
  \includegraphics[width=\textwidth]{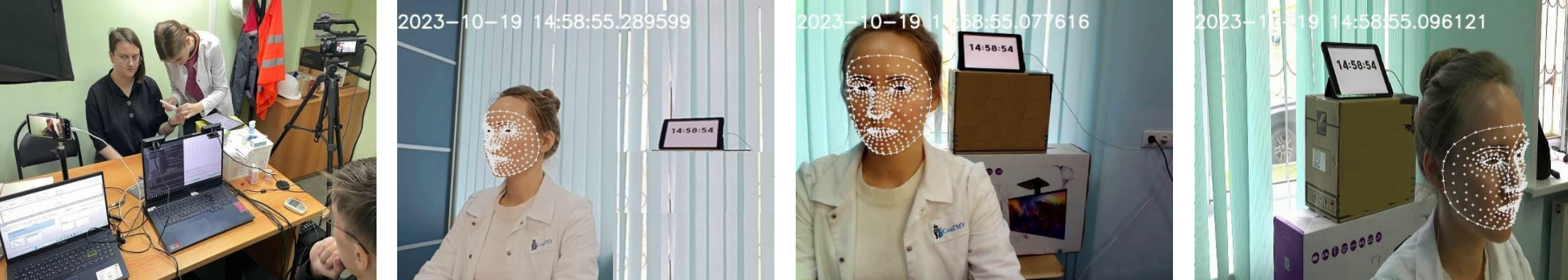}
  \caption{Our multimodal data capture setup (left) and images from three video sources with superimposed facial mesh.}
  \Description{Data acquisition setup (left) and images from three video sources with superimposed facial mesh.}
  \label{fig:teaser}
\end{teaserfigure}


\maketitle

\section{Introduction}
Pulse wave analysis is a non-invasive method widely used to evaluate cardiovascular health. Recent advancements in computer vision have enabled the estimation of pulse waves using standard video cameras and ambient light. Remote PhotoPlethysmoGraphy (rPPG) has garnered significant attention for its potential to allow background health monitoring without special medical procedures. This innovative approach is now transitioning from research to practical applications~\cite{banerjee2014heartsense}, with some technologies already integrated into everyday devices like health monitoring mirrors. These advancements facilitate the early detection of cardiovascular issues in seemingly healthy individuals, highlighting conditions such as elevated blood pressure or cardiac stiffness, which could indicate a higher risk of severe cardiovascular diseases~\cite{miotto2018reflecting}.

However, factors such as shooting conditions, lighting, and video duration can significantly impact the accuracy of health parameter estimation via rPPG. Understanding how these variables affect pulse wave extraction is crucial for algorithm training and testing. Existing datasets~\cite{SCAMPS_dataset,UBFC-Phys_dataset,MMPD_dataset} addressing this issue are often limited in size and lack comprehensive health data, such as temperature, pulse, photoplethysmogram, electrocardiogram (ECG), and other human biomarkers.

Existing datasets usually contain a medium number of subjects (10-140) recorded in a laboratory or office setting, see Table~\ref{tab:open_datasets}. Only some contain other health biomarkers, like blood pressure~\cite{iBVP_dataset}, or stress levels~\cite{UBFC-Phys_dataset}. Moreover, most known datasets are subject to restricted access and require submitting a request without a guarantee of approval, often due to privacy concerns, proprietary rights, or institutional policies. This limitation impedes collaborative research and slows down the pace of innovation. 

Our work aims to advance remote health monitoring by providing a novel publicly available dataset to streamline data handling and model training for further research. Dataset is available on Huggingface platform:

\href{https://huggingface.co/datasets/kyegorov/mcd_rppg}{\textbf{https://huggingface.co/datasets/kyegorov/mcd\_rppg}}

Additionally, we provide scripts for all experiments in GitHub repository:

\href{https://github.com/ksyegorov/mcd_rppg}{\textbf{https://github.com/ksyegorov/mcd\_rppg}}

In particular, our main contributions can be summarized as follows:
\begin{enumerate}
\item \textbf{Comprehensive Dataset}: We introduce the Multi-Camera Dataset for rPPG (MCD-rPPG), a large, diverse, open dataset designed to advance deep learning methods for pulse wave detection and human biomarker estimation. Featuring 600 subjects of varying genders and ages, the dataset includes three-minute multi-view videos recorded in resting states and after physical activity, synchronized with PPG signals, and enriched with 13 additional health biomarkers such as arterial blood pressure, ECG, and heart rate. 
\item \textbf{Fast Baseline Model}: We introduce an efficient multi-task neural network that estimates pulse waves and other health biomarkers from facial video. The model operates in real-time on a CPU, achieving a speed improvement of up to 13\% over leading models, while maintaining competitive accuracy compared to state-of-the-art approaches, even on mobile devices.
\item \textbf{Benchmarking and Comparison}: We thoroughly compare our rPPG model with state-of-the-art approaches, test its generalization capabilities in cross-dataset evaluation, and provide a baseline and benchmark for novel tasks, such as estimating various health biomarkers from facial videos.
\end{enumerate}

\begin{table}[!ht]
  \centering
  \caption{Existing open rPPG datasets}
  \footnotesize 
  \begin{tabular}{l|c|c|c|c}
  \toprule
    name & year & subjects & open & link \\ 
  \midrule
    PURE & 2014 & 10 & No & \cite{PURE_dataset} \\ 
    BP4D+ & 2016 & 140 & No & \cite{PB4D_dataset} \\ 
    COHFACE & 2017 & 40 & No & \cite{COHFACE_dataset} \\ 
    LGI-PPGI & 2018 & 25 & Yes & \cite{LGI-PPGI_dataset} \\
    UBFC-rPPG & 2019 & 42 & No & \cite{UBFC-RPPG_dataset} \\ 
    UBFC-Phys & 2021 & 56 & No & \cite{UBFC-Phys_dataset} \\ 
    SCAMPS & 2022 & 2800 (synthetic) & Yes & \cite{SCAMPS_dataset} \\
    MMPD & 2023 & 33 & No & \cite{MMPD_dataset} \\ 
    VitalVideos & 2023 & 900 & No & \cite{VitalVideos_dataset} \\ 
    iBVP & 2024 & 33 & No & \cite{iBVP_dataset} \\ 
    \bf MCD-rPPG & 2024 & 600 & Yes & Ours \\ 
  \bottomrule
  
  \end{tabular}

  \label{tab:open_datasets}
\end{table}

\section{MCD-rPPG dataset}
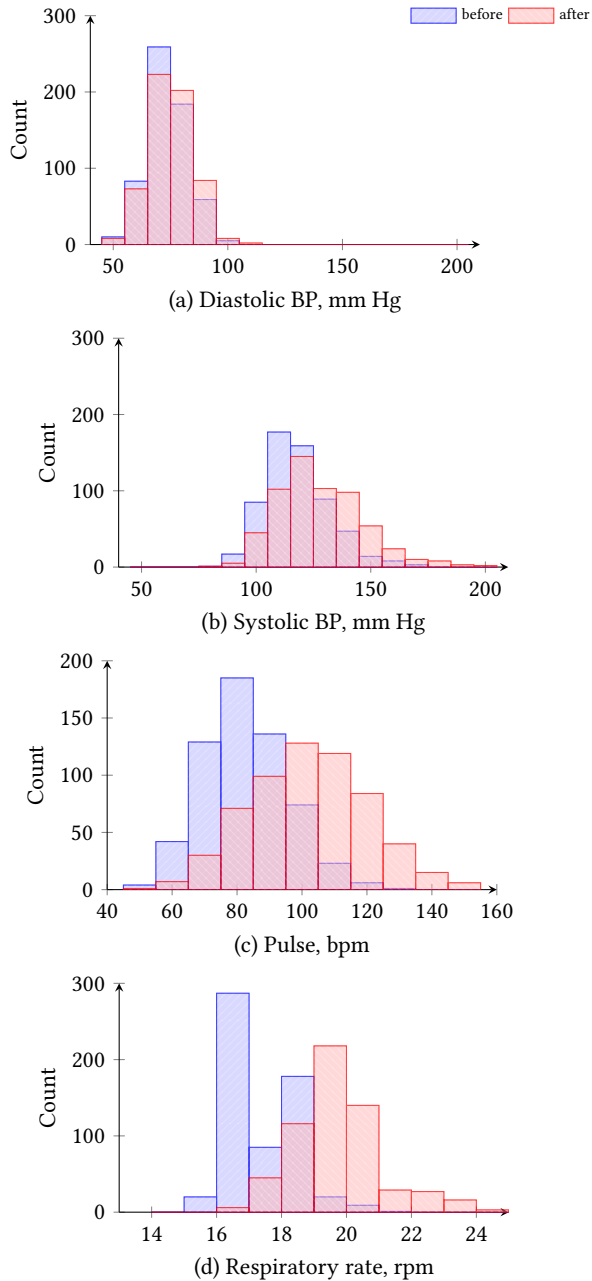
\begin{figure}[htb]
    \centering
    \begin{minipage}{0.4\textwidth}
        \centering
        \begin{tikzpicture}
            \begin{axis}[
            ybar,
            axis x line=bottom,
            axis y line=left,
            xmin=40,
            xmax=210,
            ymin=0,
            ymax=300,
            xlabel={(a) Diastolic BP, mm Hg},
            ylabel={Count},
            bar shift=0,
            width=0.95\textwidth,
            height=0.65\textwidth,
            legend entries={before, after},
            legend style={
                at={(0.8,1)}, 
                anchor=west,
                legend columns=-1,
                font=\scriptsize, 
                draw=none, 
                fill=white, 
                fill opacity=0, 
                text opacity=1, 
            },
        ]
            \addplot+[bar width=10, opacity=0.5, area legend, postaction={pattern=north east lines, pattern color=white}] coordinates {
                (50, 10) (60, 83) (70, 259) (80, 184) (90, 59) (100, 5) (110, 0) (120, 0) (130, 0) (140, 0) (150, 0) (160, 0) (170, 0) (180, 0) (190, 0) (200, 0)
            };
            \addplot+[bar width=10, opacity=0.5, area legend, postaction={pattern=north west lines, pattern color=white}] coordinates {
                (50, 8) (60, 73) (70, 223) (80, 202) (90, 84) (100, 8) (110, 2) (120, 0) (130, 0) (140, 0) (150, 0) (160, 0) (170, 0) (180, 0) (190, 0) (200, 0)
            };
        \end{axis}
        \end{tikzpicture}

        \vfill

        \centering
        \begin{tikzpicture}
            \begin{axis}[
                ybar,
                axis x line=bottom,
                axis y line=left,
                xmin=40,
                xmax=210,
                ymin=0,
                ymax=300,
                xlabel={(b) Systolic BP, mm Hg},
                ylabel={Count},
                bar shift=0,
                width=0.95\textwidth,
                height=0.65\textwidth
            ]
                \addplot+[bar width=10, opacity=0.5, area legend, postaction={pattern=north east lines, pattern color=white}] coordinates { (50, 0) (60, 0) (70, 0) (80, 1) (90, 17) (100, 85) (110, 177) (120, 159) (130, 89) (140, 47) (150, 14) (160, 8) (170, 3) (180, 0) (190, 0) (200, 0) };
                \addplot+[bar width=10, opacity=0.5, area legend, postaction={pattern=north west lines, pattern color=white}] coordinates { (50, 0) (60, 0) (70, 0) (80, 1) (90, 5) (100, 45) (110, 102) (120, 145) (130, 103) (140, 98) (150, 54) (160, 24) (170, 10) (180, 8) (190, 3) (200, 2) };
            \end{axis}
        \end{tikzpicture}

    \end{minipage}
    \hfill
    \begin{minipage}{0.4\textwidth}
        \centering
        \begin{tikzpicture}
            \begin{axis}[
                ybar,
                axis x line=bottom,
                axis y line=left,
                xmin=40,
                xmax=160,
                ymin=0,
                ymax=200,
                xlabel={(c) Pulse, bpm},
                ylabel={Count},
                bar shift=0,
                width=0.95\textwidth,
                height=0.65\textwidth
            ]
                \addplot+[bar width=10, opacity=0.5, area legend, postaction={pattern=north east lines, pattern color=white}] coordinates {
                    (50, 4) (60, 42) (70, 129) (80, 185) (90, 136) (100, 74) (110, 23) (120, 6) (130, 1) (140, 0) (150, 0)
                };
                \addplot+[bar width=10, opacity=0.5, area legend, postaction={pattern=north west lines, pattern color=white}] coordinates {
                    (50, 1) (60, 7) (70, 30) (80, 71) (90, 99) (100, 128) (110, 119) (120, 84) (130, 40) (140, 15) (150, 6)
                };
            \end{axis}
        \end{tikzpicture}

        \vfill

        \centering
        \begin{tikzpicture}
            \begin{axis}[
                ybar,
                axis x line=bottom,
                axis y line=left,
                xmin=13,
                xmax=25,
                ymin=0,
                ymax=300,
                xlabel={(d) Respiratory rate, rpm},
                ylabel={Count},
                bar shift=0,
                width=0.95\textwidth,
                height=0.65\textwidth
            ]
                \addplot+[bar width=1, opacity=0.5, area legend, postaction={pattern=north east lines, pattern color=white}] coordinates {
                    (14.5, 0) (15.5, 20) (16.5, 287) (17.5, 85) (18.5, 178) (19.5, 20) (20.5, 9) (21.5, 1) (22.5, 0) (23.5, 0) (24.5, 0)
                };
                \addplot+[bar width=1, opacity=0.5, area legend, postaction={pattern=north west lines, pattern color=white}] coordinates {
                    (14.5, 0) (15.5, 0) (16.5, 6) (17.5, 45) (18.5, 116) (19.5, 218) (20.5, 140) (21.5, 29) (22.5, 27) (23.5, 16) (24.5, 3)
                };
            \end{axis}
        \end{tikzpicture}

    \end{minipage}
    \caption{Value distribution of diastolic blood pressure~(a), systolic blood pressure~(b), pulse~(c) and respiratory rate~(d), before and after physical activity.}
    \label{img:params_before_after}
\end{figure}

Our dataset was created with 600 subjects by simultaneously measuring PPG and ECG signals using medical-grade devices (Eldar and AXMA HemoCard-BT) directly from the subject and recording videos from three different angles using various recording devices (mobile phone, video camera, and webcam). 
For each subject, two recordings were made: one in the resting state and another after light physical activity (15 squats in 30 seconds). This approach enhances the resulting dataset, which can be used to train more stable models concerning the subject's physical state. Differences between states can be illustrated by differences in pulse, blood pressure, and respiratory rate distributions, as shown in Fig.~\ref{img:params_before_after}.
A similar logic was applied when deciding whether to include different viewing angles.

Each recording session lasted approximately 3 minutes, during which the video was recorded at standard VGA resolution (\(640\times480\)) with a frame rate of 24 or 30, depending on the device. The PPG signal was sampled at a rate of 100 Hz. Additional anthropometric and biometric data were gathered before the session, as shown in Table~\ref{tab:basic_anthropometric}, using appropriate medical equipment when necessary (scales REKAM BS 630FT, thermometer Schwabe F01, tonometer AND UA-911BT-C, pulse oximeter Beurer PO~60, blood analyzers EasyTouch and EasyTouch 2, volumetric sphygmography system BPLab Angio). The stress assessment was conducted using the PSM-25 psychological stress scale, measured through a questionnaire. 

\begin{table}
\centering
\caption{Distribution of anthropometric parameters and biomarkers in the dataset}
\footnotesize 
\begin{tabular}{l|ccccccccc}
\toprule
 & mean & std & min & max \\
\midrule
Weight, kg & 65.92 & 15.79 & 43.00 & 168.00 \\
Height, cm & 169.75 & 8.87 & 147.00 & 201.00 \\
BMI, kg/m\textsuperscript{2} & 22.73 & 4.34 & 15.39 & 47.03 \\
Age, years & 23.08 & 10.90 & 18.00 & 83.00 \\
\midrule
Systolic pressure, mm Hg & 122.45 & 17.43 & 80.00 & 202.00 \\
Diastolic pressure, mm Hg & 73.79 & 9.25 & 50.00 & 108.00 \\
Saturation, \% & 98.01 & 1.29 & 86.00 & 99.00 \\
Temperature, \(^{\circ}\) & 36.56 & 0.13 & 36.00 & 37.50 \\
Hemoglobin, g/dL & 13.59 & 1.66 & 8.10 & 17.30 \\
Glycated hemoglobin, \% & 5.52 & 0.69 & 3.40 & 13.02 \\
Cholesterol, mmol/L & 4.16 & 0.83 & 0.90 & 8.00 \\
Respiratory rate, rpm & 18.05 & 1.71 & 15.00 & 24.00 \\
Heart rate, bpm & 91.93 & 18.37 & 49.00 & 153.00 \\
Arterial stiffness & 8.99 & 3.04 & 1.75 & 34.02 \\
Stress (PSM-25) & 3.04 & 1.46 & 1.00 & 7.52 \\
\bottomrule
\end{tabular}
\label{tab:basic_anthropometric}
\end{table}

\subsection{Data collection procedure}

The data collection process was organized as follows. 
At first, the subjects filled out the consent form to participate in the study with medical intervention and the consent form to process data. Then, they were asked to fill out a Psychological Stress Scale PSM-25 (Lemyr-Tessier-Fillion) by indicating the assessment of statements on an 8-point scale.

Next, the subject proceeded to the medical parameter collection stand (Fig.~\ref{fig:teaser}), where the main physiological parameters were assessed (body weight, height, temperature, blood pressure, heart rate (HR), oxygen saturation, ECG, blood glucose, glycated hemoglobin, total cholesterol, respiratory rate, and arterial wall stiffness). 
Body weight was measured using the REKAM BS 630FT floor scale. 
Blood pressure and HR were measured using an automatic tonometer AND UA-911BT-C. ECG was recorded by registering at least three channels of a 12-channel ECG using the ACSMA GemoCard-BT device. Oxygen saturation was determined by indirect oximetry of the index finger using the Beurer PO 60 pulse oximeter. 
All these devices can remotely transmit data to the database via Bluetooth. The blood glucose, total cholesterol, and glycated hemoglobin levels were determined by puncturing the index finger pad with a disposable sterile scarifier, then using test strips and blood analyzers EasyTouch (glycated hemoglobin, total cholesterol) and EasyTouch 2 (glucose). Body temperature was measured using a non-contact thermometer. Arterial wall stiffness was assessed using the Eldar photoplethysmograph or the system for volumetric sphygmography - BPLab Angio. A physician measured the respiratory rate by counting the number of respiratory movements per minute.

Immediately after that, the subject proceeded to the video and photoplethysmography data collection stand, which was a laptop with three video cameras (mobile phone Samsung A3s, video camera Sony FDR-AX43A, and webcam Defender G-lens 2599) and a photoplethysmograph connected. The subject sat down in front of the cameras, put the photoplethysmograph probe on their finger, and waited motionlessly for three minutes. An electronic clock with a second value was placed behind the subject to synchronize the video. The operator's task was to accompany the subject at this stand and monitor the lighting and validity of the collected data. After three minutes of recording, the subject performed a physical exercise of 15 squats in 30 seconds, and then all parameter recordings were repeated.

Specialized software was developed to record video and anthropometric and biometric parameters. The software operates as follows. In the user interface, the operator enters the subject's data and the recording stage (initial or after physical exercises). After that, input data is automatically validated, and the initialization process for video and PPG capturing begins. This process is performed until the internal timer reaches a predefined number of seconds. Then, the operator displays the recorded PPG signal on the screen for visual control. Finally, the program reinitializes for the next recording.

\section{Synchronization challenges}
Precise time synchronization of data is essential, as it can significantly impact the achievable metrics for training AI models.

\subsection{Video synchronization}
Although time synchronization of video streams from different recording devices is assisted to some extent by software developed for this purpose, timestamps are assigned on the computer side, which does not consider the latency introduced by the recording device and the data transfer to the computer.
Thus, we have enriched the data with an additional synchronization channel. As seen in Fig.~\ref{fig:teaser}, there is a tablet with a digital clock in the background, visible from all the cameras.
There are two major things to take into account:
\begin{itemize}
\item Firstly, the tablet clock is not perfectly synchronized with the computer, which does not prevent us from calculating time shifts between video sources pairwise;
\item Secondly, tablet clock lacks sub-second time resolution, which timestamps have.
\end{itemize}
The latter issue is solved by employing the following algorithm:
\begin{itemize}
\item Tablet clock data is extracted from each frame using optical character recognition (EasyOCR) and cleansed;
\item For each adjacent pair of frames with different times displayed on the tablet clock, the time shift is calculated as the difference between the time displayed in the last frame and half of the sum of the frame timestamps;
\item The record time shift is obtained by taking the average of all the shifts calculated in the previous step;
\item The quality of video stream synchronization is estimated for each pair of video recording devices by calculating the direct difference between their corresponding time shifts.
\end{itemize}

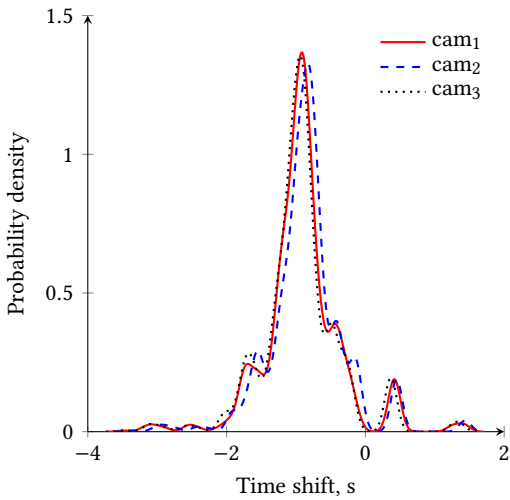
\begin{figure}[htb]
\centering
\begin{tikzpicture}
\begin{axis}[smooth, axis x line=bottom, axis y line=left, xmin=-4, xmax=2, ymin=0, ymax=1.5, xlabel={Time shift, s}, ylabel={Probability density}, legend entries={$\textrm{cam}_1$, $\textrm{cam}_2$, $\textrm{cam}_3$}, width=0.4\textwidth, height=0.4\textwidth, legend style={draw=none}, cycle multiindex* list={linestyles \nextlist color list \nextlist}]
\addplot+[mark=none, thick] table [x=xx0, y=yy0, col sep=comma] {timeshift_kde.csv};
\addplot+[mark=none, thick] table [x=xx1, y=yy1, col sep=comma] {timeshift_kde.csv};
\addplot+[mark=none, thick] table [x=xx2, y=yy2, col sep=comma] {timeshift_kde.csv};
\end{axis}
\end{tikzpicture}
\caption{Distribution of record time shifts (between frame timestamps and physical clock) estimated using KDE ($\textrm{cam}_1$ is IriunWebcam, $\textrm{cam}_2$ is FullHDwebcam and $\textrm{cam}_3$ is USBVideo).}
\label{img:timeshift_kde}
\end{figure}

The results for record time shifts (excluding 226 records, or approximately 6.3\% of the total number of records for which OCR failed) are shown in Fig.~\ref{img:timeshift_kde}. Here, the distributions for different cameras match quite closely, suggesting that the absolute value is determined by the drift of the tablet's internal clock. Additionally, it can be shown that the record time shifts for different video recording devices are linearly dependent.

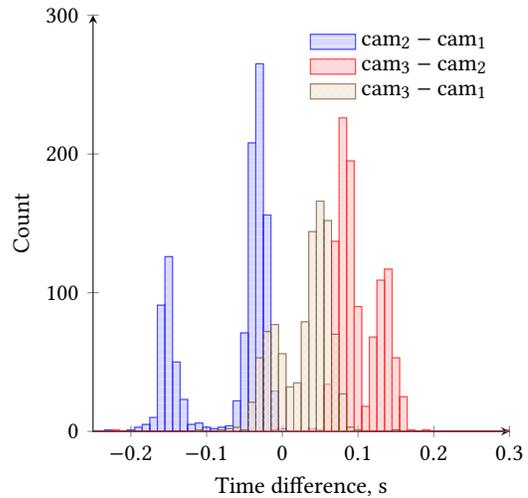
\begin{figure}[htb]
\centering
\begin{tikzpicture}
\begin{axis}[ybar, axis x line=bottom, axis y line=left, xmin=-0.25, xmax=0.3, ymin=0, ymax=300, xlabel={Time difference, s}, ylabel={Count}, bar shift=0, legend entries={$\textrm{cam}_2-\textrm{cam}_1$, $\textrm{cam}_3-\textrm{cam}_2$, $\textrm{cam}_3-\textrm{cam}_1$}, width=0.4\textwidth, height=0.4\textwidth, legend style={draw=none}]
\addplot+ [bar width=0.01, opacity=0.5, area legend, postaction={pattern=horizontal lines, pattern color=white}] table [x=grid, y=y01, col sep=comma] {timedelta_intercam.csv};
\addplot+ [bar width=0.01, opacity=0.5, area legend, postaction={pattern=crosshatch dots, pattern color=white}] table [x=grid, y=y12, col sep=comma] {timedelta_intercam.csv};
\addplot+ [bar width=0.01, opacity=0.5, area legend, postaction={pattern=crosshatch, pattern color=white}] table [x=grid, y=y02, col sep=comma] {timedelta_intercam.csv};
\end{axis}
\end{tikzpicture}
\caption{Distribution of time shift between different video sources ($\textrm{cam}_1$ is IriunWebcam, $\textrm{cam}_2$ is FullHDwebcam and $\textrm{cam}_3$ is USBVideo).}
\label{img:timedelta_intercam}
\end{figure}

\begin{figure}[t]
\centering
\begin{tikzpicture}
\begin{axis}[ybar, axis x line=bottom, axis y line=left, xmin=-20, xmax=20, ymin=0, ymax=310, xlabel={Optimal shift, frames}, ylabel={Count}, bar shift=0, legend entries={$\textrm{cam}_1$, $\textrm{cam}_2$, $\textrm{cam}_3$}, width=0.4\textwidth, height=0.4\textwidth, legend style={draw=none}]
\addplot+ [bar width=1, opacity=0.5, area legend, postaction={pattern=horizontal lines, pattern color=white}] table [x=grid, y=y1, col sep=comma] {framedelta_ppg.csv};
\addplot+ [bar width=1, opacity=0.5, area legend, postaction={pattern=crosshatch dots, pattern color=white}] table [x=grid, y=y2, col sep=comma] {framedelta_ppg.csv};
\addplot+ [bar width=1, opacity=0.5, area legend, postaction={pattern=crosshatch, pattern color=white}] table [x=grid, y=y3, col sep=comma] {framedelta_ppg.csv};
\end{axis}
\end{tikzpicture}
\caption{Distribution of time shift between ground truth PPG and reconstructed PPG ($\textrm{cam}_1$ is IriunWebcam, $\textrm{cam}_2$ is FullHDwebcam and $\textrm{cam}_3$ is USBVideo).}
\label{img:framedelta_ppg}
\end{figure}
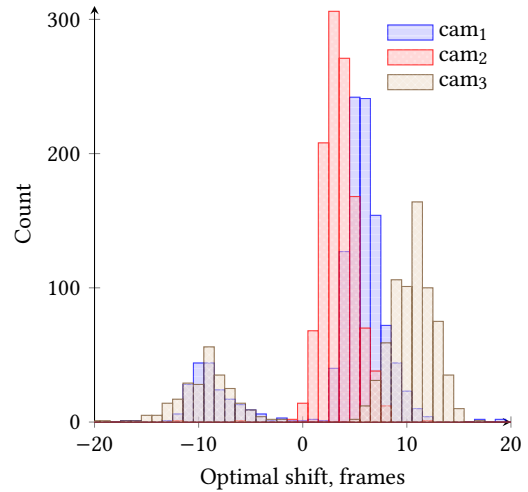

Fig.~\ref{img:timedelta_intercam} shows that the time difference is generally within ~\(\pm0.2\) seconds, constituting only a fraction of the average heartbeat cycle length. Moreover, the proposed method can further refine the temporal data.

\subsection{PPG synchronization}
Considering possible rPPG applications, another important point to consider is the synchronization of video and PPG data. To estimate it, we compared ground truth PPG signal with PPG signals reconstructed from video using the POS algorithm as follows:
\begin{itemize}
\item all signals are filtered using 4-order Chebyshev Type~II band-pass filter with cutoff frequencies of 0.4~Hz and 8~Hz;
\item Discrete shift was determined in terms of the optimization problem for maximization of the Pearson correlation coefficient between reconstructed signals and ground truth.
\end{itemize}

\begin{figure*}[t]
 \centering
 \includegraphics[width=\textwidth]{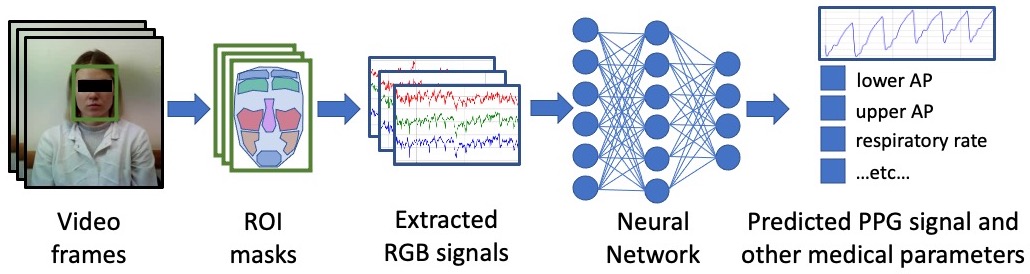}
 \caption{Overview of our baseline model.}
 \label{fig:model_description}
\end{figure*}

The obtained results are shown in Fig.~\ref{img:framedelta_ppg} (326 records, or approximately 9.1\% of the total number of records for which the POS algorithm failed, were excluded). The second set of peaks on the left side is most probably caused by a time shift approaching half a period of the PPG signal, in which case, the determination of the nearest maximum becomes ambiguous.
Considering that the frontal facing camera ($\textrm{cam}_2$) shows significantly better synchronization, the subpar results for other cameras are at least partially attributed to the shortcomings of the POS algorithms.







\section{Baseline Model}

Modern rPPG models can be either unsupervised or supervised. Unsupervised methods can be divided into two subgroups: methods based on reading micro-movements (Ballistocardiographic)~\cite{mit_coordinates} and methods based on micro changes in the color of facial pixels~ \cite{POS_paper,CHROME_paper}. Supervised methods, such as \cite{PhysFormer_paper,PhysNet_paper,DeepPhys_paper}, are based on deep learning and let the neural network determine where and what to look at. Such methods are more accurate and better predict the PPG shape, but require datasets for training and often are subject to overfitting. This can be seen in the results of \cite{rppg_toolbox}, where unsupervised POS~\cite{POS_paper} and OMI~\cite{OMIT_paper} models showed results that were more robust across datasets compared to supervised approaches, as well as in \cite{remote_biosensing}, where all supervised models showed a significant drop in quality during cross-dataset evaluation.

In this regard, we decided to take a hybrid approach relying on domain-specific pre-processing, followed by processing with a specialized neural network adapter, similar to~\cite{liu2021efficientphys}. Because the network has fewer parameters, it is much less susceptible to overfitting while maintaining high quality and the ability for fine-tuning and domain adaptation, if necessary. Additionally, adding new targets like arterial pressure or respiratory rate to neural networks is possible without significantly altering their architecture.


For a domain-specific pre-processing tool, we used ideas from \cite{CHROME_paper,POS_paper}. It is essential to consider the anatomical properties of the blood supply to the face~\cite{face_blood_flow} and detect those areas of the face with the greatest influence of the pulse wave, significantly reducing the noise level in the signal. We select the face region-of-interest (ROI) using a technique from~\cite{rppg_roi_selection}, allowing the neural network to choose the signal of different ROIs in the required proportions.



The final model architecture is shown in Fig.~\ref{fig:model_description}. First, we detect a face and highlight the ROI using the FaceMesh model from the mediapipe~\cite{mediapipe_paper}. We then select a set of multiple regions and compute the mean pixel value in each ROI for each frame. The received signals are fed into a neural network, producing a PPG signal and medical parameter predictions. The model is a fully convolutional 1-dimensional feature pyramid network \cite{lin2017feature}. It allows the neural networks to operate on different lengths of the input signal without slicing it in a sliding window manner. The resulting pipeline is very fast and surpasses even unsupervised models in inference time on GPU and CPU. The proposed model is a good foundation for analyzing an extended set of biomarkers. While the ROI-based approach is indeed common in rPPG, our main contribution is developing a blazing-fast model capable of working on small devices like phones and wearables while being on par with high-capacity models in terms of accuracy.

\section{Experiments}

\begin{table*}[t]
\centering
\caption{Comparison of performance of different models (MAE). With \textbf{bold} font, we highlight the best-performing model on each test dataset. With \underline{underline} font, we highlight the best-performing model, which is not trained on the test dataset in question. The last line describes the performance of our model trained on the MCD-rPPG dataset.}
\scriptsize 
  \resizebox{\textwidth}{!}{
  \begin{tabular}{l|l|c|c|c|c|c|c|c|c}
  \toprule
     & Train & MCD-rPPG & MCD-rPPG & MMPD & MMPD & SCAMPS & SCAMPS & UBFC-rPPG & UBFC-rPPG \\ 
    Model   &  dataset  & PPG (Ours)   & HR (Ours)   & PPG  & HR  & PPG  & HR & PPG & HR \\
  \midrule
    PBV\cite{pbv_paper} & - & 0.85 & 15.37 & 0.80 & 37.11 & 0.88 & 35.93 & 0.86 & 30.27 \\ \hline
    OMIT\cite{OMIT_paper} & - & 0.80 & 4.78   & 0.77    & 15.33      & 0.86    & 16.27     & 0.84    & 1.95 \\ \hline
    POS\cite{POS_paper} & - & 0.87 &  \underline{3.80} & 1.08 & \underline{15.36} & 1.41 & \underline{16.02} & 1.52 & \underline{\textbf{1.17}} \\ \hline
     PhysFormer\cite{PhysFormer_paper} & MMPD & 0.89±0.02 & 13.57±1.40 & 0.82±0.02 & 22.67±1.21 & 0.90±0.02 & 28.45±0.75 & 0.79±0.04 & 8.79±5.96 \\ 
      & SCAMPS & 1.10±0.00 & 46.38±0.00 & 0.97±0.00 & 30.01±0.00 & 0.13±0.00 & 0.40±0.00 & \underline{\textbf{0.67±0.00}} & \underline{\textbf{1.17±0.00}} \\ 
      & UBFC-rPPG & \underline{0.81±0.01} & 43.65±3.71 & \underline{\textbf{0.76±0.01}} & 42.09±7.21 & \underline{0.84±0.02} & 49.77±4.70 & 0.77±0.01 & 37.40±0.98 \\ 
      & \bf MCD-rPPG & 0.46±0.01 & 4.08±0.12 & 0.98±0.03 & 22.61±0.51 & 1.08±0.03 & 23.33±1.29 & 1.30±0.06 & 1.27±1.51 \\ \hline
     iBVPNet\cite{iBVP_dataset} & MMPD & 0.99±0.02 & 36.28±6.10 & 0.87±0.01 & 21.02±0.20 & 0.96±0.01 & 34.60±2.04 & 0.93±0.01 & 21.63±5.16 \\ 
      & SCAMPS & 1.07±0.01 & 59.48±1.93 & 1.01±0.00 & 24.19±0.67 & 0.54±0.00 & 1.63±0.08 & 0.84±0.03 & 7.91±2.62 \\ 
      & UBFC-rPPG & 0.96±0.03 & 35.31±7.24 & 0.86±0.01 & 23.53±3.92 & 0.90±0.02 & 31.36±2.17 & 0.85±0.01 & 7.47±2.21 \\ 
      & \bf MCD-rPPG & 0.68±0.01 & 4.83±0.44 & 1.01±0.01 & 17.12±0.51 & 1.05±0.00 & 25.74±0.23 & 1.21±0.01 & 4.69±0.71 \\ \hline
         RhythmFormer\cite{rhythmformer} & MMPD & 0.94±0.02 & 17.54±4.30 & 0.77±0.00 & 17.28±2.21 & 0.91±0.01 & 30.34±1.00 & 0.85±0.01 & 11.52±2.80 \\ 
      & SCAMPS & 1.05±0.01 & 58.58±8.14 & 1.00±0.01 & 26.08±0.44 & \textbf{0.08±0.00} & \textbf{0.20±0.10} & 0.75±0.03 & 10.25±2.68 \\ 
      & UBFC-rPPG & 0.87±0.01 & 19.14±1.93 & 0.82±0.01 & 24.81±0.69 & 0.99±0.01 & 28.53±0.43 & 0.86±0.02 & 21.53±3.29 \\ 
      & \bf MCD-rPPG & \textbf{0.43±0.00} & \textbf{2.82±0.13} & 0.98±0.01 & 16.63±0.96 & 1.06±0.02 & 21.45±1.65 & 1.46±0.02 & 2.39±0.35 \\ \hline
     Ours (Fig.~\ref{fig:model_description}) & MMPD & 1.32±0.02 &7.58±1.3 & 0.94±0.01 & \textbf{15.26±0.49} & 1.16±0.02 & 27.77±1.75 & 0.96±0.12 & 2.39±0.46 \\
      & SCAMPS & 1.20±0.01 & 41.17±4.30 & 1.06±0.03 & 34.44±3.81 & 0.50±0.04 & 10.60±0.77 & 0.87±0.03 & 13.62±3.57 \\ 
      & UBFC-rPPG & 1.17±0.05 & 23.09±6.67 & 1.01±0.02 & 27.93±3.41 & 0.92±0.08 & 37.46±0.62 & 0.77±0.02 & 6.84±5.27 \\  
      & \bf MCD-rPPG & 0.68±0.03 & 4.86±0.36 & 1.20±0.02 & 17.53±0.68 & 1.27±0.05 & 20.20±1.39 & 1.42±0.08 & 4.13±0.92 \\
    \bottomrule
  \end{tabular}
  }
\label{tab:model_results}
\end{table*}

\begin{table}
\centering
\caption{Performance for different camera views}
\footnotesize 
\begin{tabular}{l|ccc|ccccccc}
\toprule
 & \multicolumn{3}{c|}{Speed of Inference (ms)} & \multicolumn{4}{c}{Metrics (MAE)} \\
\cmidrule(lr){2-4} \cmidrule(lr){5-8}
 & CPU & GPU & Size & Frontal & Side & Frontal & Side \\
Model & seconds & seconds & Mb & PPG & PPG & HR & HR \\
\midrule
PBV & 0.18 & 0.18 & 0 & 0.85 & 0.86 & 15.37 & 40.44 \\
OMIT & 0.17 & 0.17 & 0 & 0.80 & 1.11& 4.78 & 22.35 \\
POS & 0.26 & 0.26 & 0 & 0.87 & 1.25 & 3.80 & 16.40 \\
PhysFormer & 0.93 & 0.31 & 28.4 & 0.46 & 0.97 & 4.08 & 10.68 \\
RhythmFormer & 0.97 & 0.33 & 12.9 & 0.43 & 0.91 & 2.82 & 7.33 \\
iBVPNet & 0.93 & 0.28 & 5.5 & 0.68 & 0.99 & 4.83 & 11.42 \\
Ours & \textbf{0.15} & \textbf{0.16} & 3.9 & 0.68 & 1.10 & 4.86 & 14.01 \\
\bottomrule
\end{tabular}
\label{tab:combined_results}
\end{table}

\begin{table}
\centering
\caption{Metric for predicting biomarkers compared to a naive baseline (all metrics are MAE, except for Sex)}
\footnotesize 
\begin{tabular}{l|cc}
\toprule
Target & Baseline & Model \\ 
\midrule
Systolic pressure, mm Hg & 13.78 & \textbf{12.82} \\ 
Diastolic pressure, mm Hg & \textbf{7.50} & 8.39 \\
Glycated hemoglobin, \% & 0.43 & \textbf{0.41} \\ 
Cholesterol, mmol/L & 0.66 & \textbf{0.60} \\ 
Respiratory rate, rpm & 1.36 & \textbf{1.20} \\ 
Arterial stiffness & 2.19 & \textbf{2.04} \\
Age, years & 5.71 & \textbf{3.91} \\ 
BMI & \textbf{3.18} & 3.37 \\ 
Stress (PSM-25) & 1.20 & \textbf{1.07} \\ 
Saturation, \% & 0.98 & 0.98 \\ 
Sex (Accuracy) & 0.61 & \textbf{0.64} \\
\bottomrule
\end{tabular}
\label{tab:medical_results}
\end{table}

We trained our model by splitting videos and PPG signals into 20-second windows and feeding them to the neural network. The main target of the training was to predict the PPG signal. Still, the dataset had additional targets: systolic and diastolic arterial pressure, glycated hemoglobin, cholesterol, respiratory rate, arterial stiffness, age, sex, BMI, stress level, and saturation. All targets were normalized with a standard scaler on the training subset, and the loss function was the sum of mean squared error (MSE) losses for each target. For optimization, we utilized the Adam optimizer. We used the mean average error (MAE) of HR for each 10-second segment as our primary metric. To determine HR, we use the algorithm \cite{fft_hr_detection}. Table~\ref{tab:model_results} presents the PPG and HR estimation results. Two metrics are shown: the mean average error (MAE) of the predicted PPG signal and the MAE of HR estimation. HR is predicted by selecting the most powerful frequency of 0.5-3 Hz of the predicted PPG signal. The training procedure and all hyperparameters necessary for repeating the results can be obtained from the repository.

The result of the biomarkers estimation is presented in Table~\ref{tab:medical_results}. Our model performs better than the straightforward baseline, the optimal constant value fitted on a training subset.

One of the goals of gathering this dataset was to research the impact of different circumstances and camera parameters on the quality of remote medical scanning. For this purpose, we provide multi-view videos from three sources and our experimental results. First, as seen in Fig.~\ref{fig:teaser}, three camera views cover the face of the patient differently, so we tested the drop in quality. The results are presented in Table~\ref{tab:combined_results}. We can see the pattern where unsupervised methods, while strong in cross-dataset generalization, cannot adapt to different view angles, and neural networks generally perform better in these scenarios.

Table~\ref{tab:combined_results} also shows our model's inference time and size advantage. Table speed advantage on CPU is 13\% better than the previous best model. Time was measured by running 200 20-second video segments sequentially. However, the results in terms of model MAE are not straightforward and show multidirectional trends. This is consistent with independent benchmarks like rPPG-Toolbox~\cite{rppg_toolbox} and Remote Biosensing Benchmark~\cite{remote_biosensing}, where different models excel in different training and testing setups. We believe it is due to the limited sizes of all available datasets. This comparison confirms our success in developing small and fast models that achieved performance metrics comparable to competitive models.

\section{Conclusion}
In this paper, we introduce a unique large MCD-rPPG dataset for rPPG and health biomarkers estimation. The dataset contains 3-minute video recordings from 600 subjects of different genders and ages filmed in two states (quiet and after exercise), aligned with 13 biomarkers. A photoplethysmograph was synchronized with multiple webcams to ensure proper parameter alignment, enabling pulse wave detection from facial video recordings. Additionally, we developed a fast, lightweight multitask rPPG model (Fig.~\ref{fig:model_description}) trained on our dataset, which relies on domain-specific pre-processing.

We believe the public release of our dataset and model will accelerate progress in AI-driven medical assistants~\cite{blinov2024gigapevt} and multimedia applications (e.g., emotion-aware telemedicine, stress monitoring via video calls, or fitness tracking via smartphone cameras), which rely on robust estimation of pulse waves and health biomarkers from everyday recording devices like mobile phones and webcams.

\section*{Acknowledgments} 
The work of A. Savchenko was supported by a grant, provided by the Ministry of Economic Development of the Russian Federation in accordance with the subsidy agreement (agreement identifier 000000C313925P4G0002) and the agreement with the Ivannikov Institute for System Programming of the Russian Academy of Sciences dated June 20, 2025 No. 139-15-2025-011.

\bibliographystyle{ACM-Reference-Format}
\bibliography{sample-base}

\end{document}